\documentclass[runningheads]{llncs}
\usepackage[final,year=2026]{eccv}
\usepackage{eccvabbrv}
\usepackage{graphicx}
\usepackage{booktabs}
\usepackage{amsmath}
\usepackage{array}
\usepackage{multirow}
\usepackage{float}
\usepackage[accsupp]{axessibility}
\usepackage{hyperref}
\usepackage{orcidlink}

\begin{document}

\title{From Recovery to Drop-off: How Action Post-training Reduces a VLM's Late-Layer Depth Decodability}
\titlerunning{Action Post-training Reduces Late-Layer Depth Decodability}

\author{Alexander Hackett\inst{1,2} \and Arnaud Denis-Remillard\inst{3,2} \and Axel Cassou\inst{4}}
\authorrunning{A. Hackett et al.}
\institute{New York University \and Reflex \and Universit\'e de Montr\'eal \and Maastricht University}

\maketitle

\begin{abstract}
How much of a vision-language model's (VLM) spatial understanding remains decodable after the action post-training process of building a vision-language-action model (VLA)? We probe depth perception, a primitive of spatiogeometric understanding, from every decoder layer of a weight-matched open-source base VLM/VLA pair: Molmo2-ER and MolmoAct2-LIBERO. First, the VLA decodes depth worse at every layer, a persistent gap we call the floor. Second, the degradation is not uniform: while the base VLM's depth decodability improves through its final layers, the VLA's collapses, an additional late-layer drop we call the cliff. We causally localize the cliff to late-layer MLP interference: ablating the late-layer MLP writes recovers the majority of the terminal decodability cliff, while matched attention ablations and the same intervention in the weight-matched base VLM produce no comparable recovery. A module-level decomposition explains this dissociation: the base VLM carries depth most accessibly in accumulated MLP writes, whereas action post-training collapses depth decodability in the late accumulated writes.
\end{abstract}

\section{Introduction}
\label{sec:intro}
Vision-language-action (VLA) models are effectively pretrained vision-language models (VLMs) post-trained into embodied policies that map observations and language to actions~\cite{brohan2022rt1,brohan2023rt2,octo2024,kim2024openvla,black2024pi0}. The adaptation grafts an action objective onto the backbone, originally by discretizing actions into tokens under the same next-token loss~\cite{brohan2023rt2,kim2024openvla}, and increasingly by continuous flow matching~\cite{black2024pi0,fang2026molmoact2}. This recipe rests on the hope that the VLM's learned visual representations transfer cleanly to motor control, a hope recent evidence complicates~\cite{kachaev2025dontblind,anon2025swol}. We ask how much of the VLM's spatiogeometric understanding is preserved by action post-training, where in the network it is lost, and why. 

We answer this with a layer-wise probing study on a weight-matched pair: Molmo2-ER and MolmoAct2-LIBERO~\cite{fang2026molmoact2}. The two share the same architecture up to the action expert and are identical at initialization, differing only by action post-training, which makes a layer-for-layer comparison well-defined. We take depth perception as a primitive of spatiogeometric understanding and train a capacity-matched Dense Prediction Transformer (DPT) probe~\cite{ranftl2021dpt} on hidden states from every decoder layer, supervised by a monocular-depth teacher~\cite{lin2025depthanything3}.

We present two findings. First, MolmoAct2 decodes depth worse than its base VLM at every layer, a persistent gap we call the \emph{floor}. Second, while the base VLM's depth decodability increases measurably through its final layers, the VLA's collapses over those same layers, a collapse we call the \emph{cliff}, ultimately an inversion relative to the VLM.

\begin{figure}[H]
  \centering
  \includegraphics[width=\linewidth]{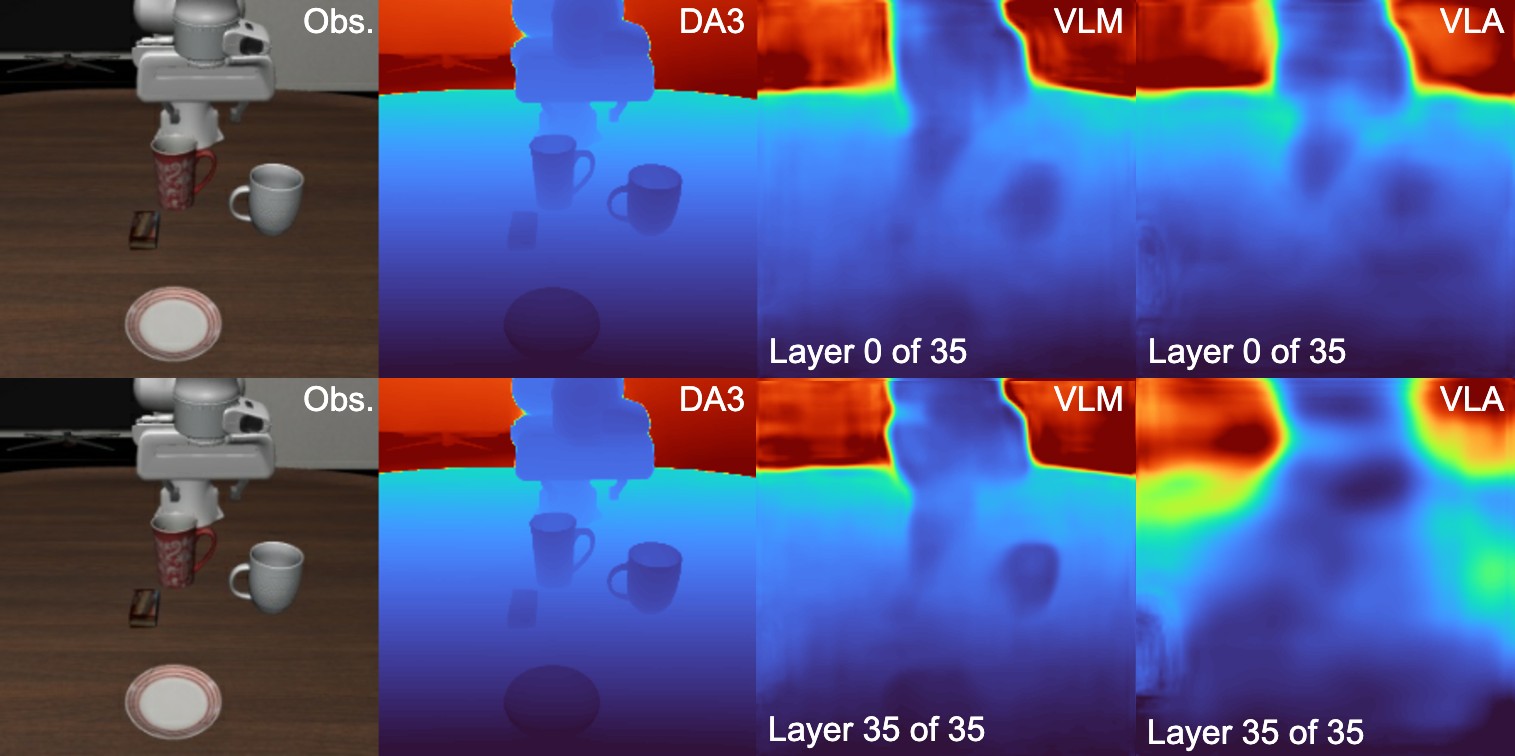}
  \caption{The cliff, qualitatively. DPT-probe depth readouts of the same LIBERO observation (Obs.; Depth-Anything-3; Molmo2-ER DPT head, MolmoAct2-LIBERO DPT head) at the first and final decoder layer. Between the first and final layer, the base VLM's readout sharpens while the VLA's collapses.}
  \label{fig:qual}
\end{figure}

The cliff, depicted qualitatively in \cref{fig:qual}, is the phenomenon we set out to explain. While a probe alone can identify what information is decodable at a given layer, it cannot isolate the specific modules or channels responsible for reading, writing, and ultimately interfering with that spatiogeometric data. In \cref{sec:causal} we distinguish these by intervention. Each layer's contribution to the residual stream is an additive write, meaning any single term can be dropped. Sweeping three-layer ablation windows over the full stack, for both modules and both models, we find that deleting the late MLP writes recovers the majority of the terminal decodability drop, whereas attention ablations, or the same intervention on the weight-matched base VLM, produce no comparable recovery. In \cref{sec:decomp} we ask why: probing accumulated MLP writes separately from the stream they sum into, we find that the base VLM holds most decodable depth information in those writes, and that action post-training collapses their decodable depth over the final blocks.

The remainder of the paper unpacks these findings: first, we map the layerwise drop in depth decodability (\cref{sec:depth}); second, we causally localize the cliff to late MLP writes (\cref{sec:causal}); and finally, we demonstrate that accumulated MLP writes carry the most depth-decodable signal before action post-training collapses it (\cref{sec:decomp}).

\section{Related Work}
\label{sec:related}
\paragraph{Representation Degradation under Action Post-Training.}
A growing line of work reports that action post-training a VLM into a VLA degrades the backbone's pretrained representations and treats the degradation as something to remedy during training~\cite{kachaev2025dontblind,anon2025swol}. Kachaev et al. ~\cite{kachaev2025dontblind} document erosion of visual-language representations and counter it with an alignment-forcing objective; concurrent work reports the same phenomenon layer-wise and likewise proposes a solution via an auxiliary temporal-consistency objective~\cite{anon2025swol}. These studies, however, probe semantic targets: ImageNet/COCO-style recognition, task classification. It is known that semantic task performance does not necessarily transfer to decodable spatiogeometric structure~\cite{bolya2025perceptionencoder,banani2024probe3d,kar2025lexicon3d}. Because existing VLA degradation analyses are usually used to motivate corrective training methods, they also stop short of explaining the trained model's internal mechanism: they do not identify which trained submodule reduces geometric decodability or test whether deleting that submodule produces a more depth-decodable representation. We ask the layerwise geometric question against a weight-matched base and localize the reduction to a specific module with a post-hoc intervention.

\paragraph{Layerwise Mechanistic Analysis in MLLMs.}
Closest in method is Wu et al.~\cite{wu2026dropoff}, which decomposes a multimodal LLM layer by layer and shows that an adapter network corrupts intermediate features which later LLM layers recover, using per-layer segmentation probes and attention knockout. Recovering the same broad drop-then-recover trajectory provides a qualitative sanity check on the layerwise probing setup and establishes the base-model reference pattern before action tuning. Our result is the inversion of that template under action post-training. Wu et al.~\cite{wu2026dropoff} studies semantic segmentation in ordinary MLLMs and attributes recovery to attention-mediated later layers; we study depth in a weight-matched VLM/VLA pair and find that late MLP computation causally reduces final-layer depth decodability. Our decomposition inherits the view that feed-forward sublayers write content into the residual stream~\cite{elhage2021mathematical,geva2021kvmemories,geva2022promoting,meng2023rome}, but contrasts with factual-recall localization: here the MLP writes are not a single fact circuit, but the accumulated feed-forward pathway by which geometric information remains accessible.

\begin{table}[htbp]
\centering
\caption{Prior works versus our contributions.}
\label{tab:prior}
\scriptsize
\begin{tabular}{lccccc}
\toprule
 & Weight-matched & Spatial & Full-stack & Submodule & Post-hoc \\
 & VLM baseline & geometry & localization & localization & recovery \\
\midrule
Don't Blind Your VLA~\cite{kachaev2025dontblind} & \checkmark & -- & -- & -- & -- \\
Drop-off to Recovery~\cite{wu2026dropoff} & -- & -- & \checkmark & -- & \checkmark \\
Ours & \checkmark & \checkmark & \checkmark & \checkmark & \checkmark \\
\bottomrule
\end{tabular}
\end{table}

\section{Preliminaries}
\label{sec:prelim}
\subsection{Vision-Language-Action Models}
Let the multimodal token sequence be
\begin{equation}
x_{1:n} = [x_{1:k}, x_{k+1:n}],
\end{equation}
where $x_{1:k}$ are visual tokens and $x_{k+1:n}$ are textual instruction tokens. They are obtained from two encoders,
\begin{equation}
x_{1:k}=E_{\mathrm{image}}(I)\in\mathbb{R}^{k\times d_e},\quad x_{k+1:n}=E_{\mathrm{text}}(c)\in\mathbb{R}^{(n-k)\times d_e}.
\end{equation}
The combined sequence is processed by a multimodal Transformer backbone $B_\theta:\mathbb{R}^{n\times d_e}\to\mathbb{R}^{n\times d_e}$ with $L$ stacked layers, indexed $i=0,\ldots,L-1$. Denoting hidden states after layer $i$ by $h^i_{1:n}\in\mathbb{R}^{n\times d_e}$, with $h^{-1}_{1:n}:=x_{1:n}$, each layer applies two submodules sequentially, each adding its output, or \emph{write}, into the stream:
\begin{align}
a^i_{1:n} &= \mathrm{Attn}^i(\mathrm{LN}(h^{i-1}_{1:n})), \\
m^i_{1:n} &= \mathrm{MLP}^i(\mathrm{LN}(h^{i-1}_{1:n}+a^i_{1:n})), \\
h^i_{1:n} &= h^{i-1}_{1:n}+a^i_{1:n}+m^i_{1:n}.
\label{eq:block}
\end{align}
Here LN denotes the block pre-normalization, RMSNorm in our pair, and the MLP reads the stream after the attention write has entered it. Unrolled, $h^\ell=x+\sum_{i=0}^{\ell}(a^i+m^i)$: the hidden state is the running sum of every module write. This is the decomposition our ablations and module-level probes use: a write can be deleted from the stream in \cref{sec:causal}, and accumulated MLP writes can be probed separately from the stream in \cref{sec:decomp}.

For an autoregressive VLM target $y_{1:m}$, the model conditions at decoding step $t$ on the concatenation $[x_{1:n},y_{1:t-1}]$ and defines
\begin{equation}
p_\theta(y_t\mid x_{1:n},y_{1:t-1})=\mathrm{softmax}(W_o h^{L-1}_{n+t-1})_{y_t},
\end{equation}
where $W_o$ projects to the token vocabulary and the causal mask ensures the readout state depends only on the input and previous targets. Training uses the standard teacher-forced next-token loss,
\begin{equation}
\mathcal{L}_{\mathrm{LM}}(\theta)=\mathbb{E}_{(x,y)\sim\mathcal{D}}\left[-\sum_{t=1}^{m}\log p_\theta(y_t\mid x_{1:n},y_{1:t-1})\right].
\end{equation}

We instantiate this setup with Molmo2-ER as the base VLM and MolmoAct2-LIBERO as its VLA counterpart~\cite{fang2026molmoact2}. The pair is weight-matched through the shared backbone up to the action expert, differing only by action post-training. The shared backbone has $L=36$ decoder layers and hidden width $d_e=2560$; visual tokens come from SigLIP2~\cite{tschannen2025siglip2}.

Action post-training proceeds in three stages. First, trajectories are discretized by a DCT--BPE action tokenizer~\cite{pertsch2025fast} and the backbone is trained under the same next-token objective as text, on a mixed robot--multimodal corpus that keeps visual tokens grounded in general imagery. Second, a flow-matching action expert is added, cross-attending to the backbone's per-layer keys and values; the backbone is frozen following knowledge insulation~\cite{driess2025insulation} and thus action gradients update only the expert. Third, the entire model is fine-tuned on LIBERO on robot data only, with the insulation released: action gradients now reach the backbone.

\subsection{Depth Probing}
We supervise probes with Depth-Anything-3~\cite{lin2025depthanything3}, a monocular depth estimator used as an image-computable teacher. At every decoder layer we train a DPT head~\cite{ranftl2021dpt} of identical capacity, following capacity-matched 3D-awareness probing~\cite{banani2024probe3d}. We report depth accuracy as $d_1$, the fraction of pixels with $\max(\hat d/d,d/\hat d)<1.25$, after affine alignment of the prediction to the teacher at evaluation.

\begin{figure}[t]
  \centering
  \includegraphics[width=.6\linewidth]{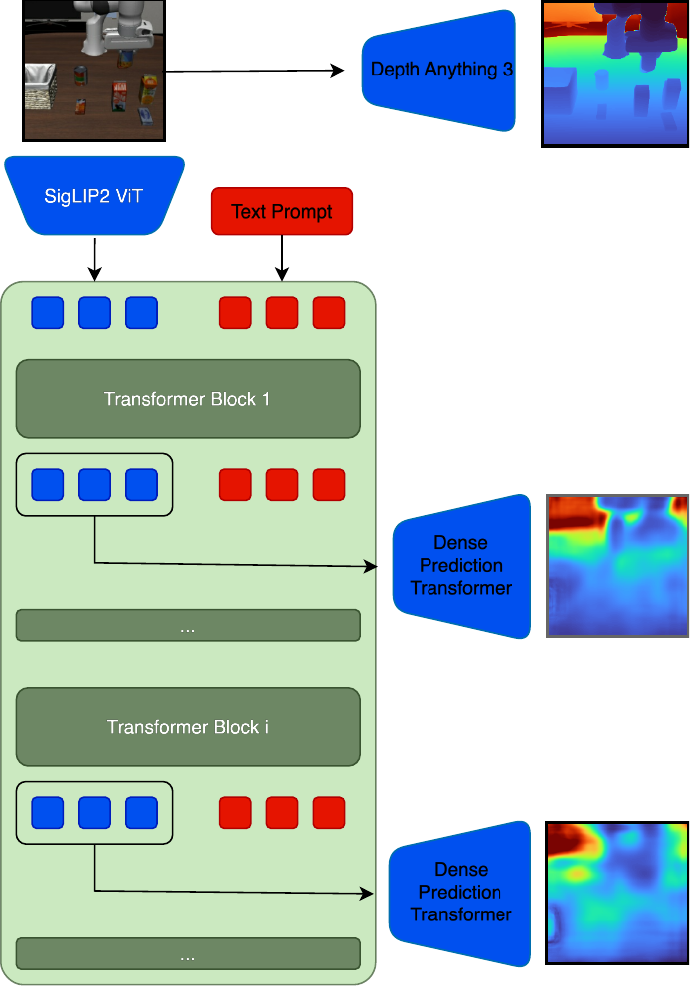}
  \caption{Dense Prediction Transformer probing schematic. A LIBERO observation is fed to the VLM/VLA backbone; a capacity-matched DPT head decodes depth from the visual tokens at every decoder layer, supervised by a Depth-Anything-3 teacher. One probe is trained per layer and per model.}
  \label{fig:dpt-probe}
\end{figure}

\subsection{Interpretability Tools}
The residual stream is the embedding plus accumulated attention and MLP writes~\cite{elhage2021mathematical}. Probing the stream separately from accumulated MLP writes $\sum_{i=0}^{\ell}m^i$ lets us ask whether depth rides the stream or the writes that feed it. Activation ablation zeroes a module's write into the residual stream and measures the downstream effect. For ridge probes we report $\mathrm{SNR}=R^2/(1-R^2)$, the classical ratio of linearly decodable to residual variance~\cite{cohen1988statistical}; here ``noise'' simply denotes target variance unexplained by the linear probe, including nonlinear encoding, superposed task-irrelevant features, teacher noise, and ordinary prediction error. We use ridge scores only as within-model corroboration: cross-model linear decodability conflates how much depth is represented with how linearly it is formatted, and action tuning may alter both.

\section{Experiments}
\subsection{Depth Probing}
\label{sec:depth}
\paragraph{Protocol.}
We read depth from hidden states of the weight-matched pair with a probe held at fixed capacity across every layer. Fig. \ref{fig:dpt-probe} depicts our probing methodology. For each decoder layer $\ell\in\{0,\ldots,35\}$ of each model, we read residual-stream hidden states at visual-token positions from LIBERO \cite{liu2023libero} frames resized to 256 pixels across both camera views, and train an identical DPT head supervised by Depth-Anything-3. We split by rollout, so near-duplicate frames from one episode never straddle train and validation, and report best-validation $d_1$ after affine alignment. Additional training and implementation details are provided in the supplementary material.

\begin{figure}[t]
  \centering
  \includegraphics[width=.78\linewidth]{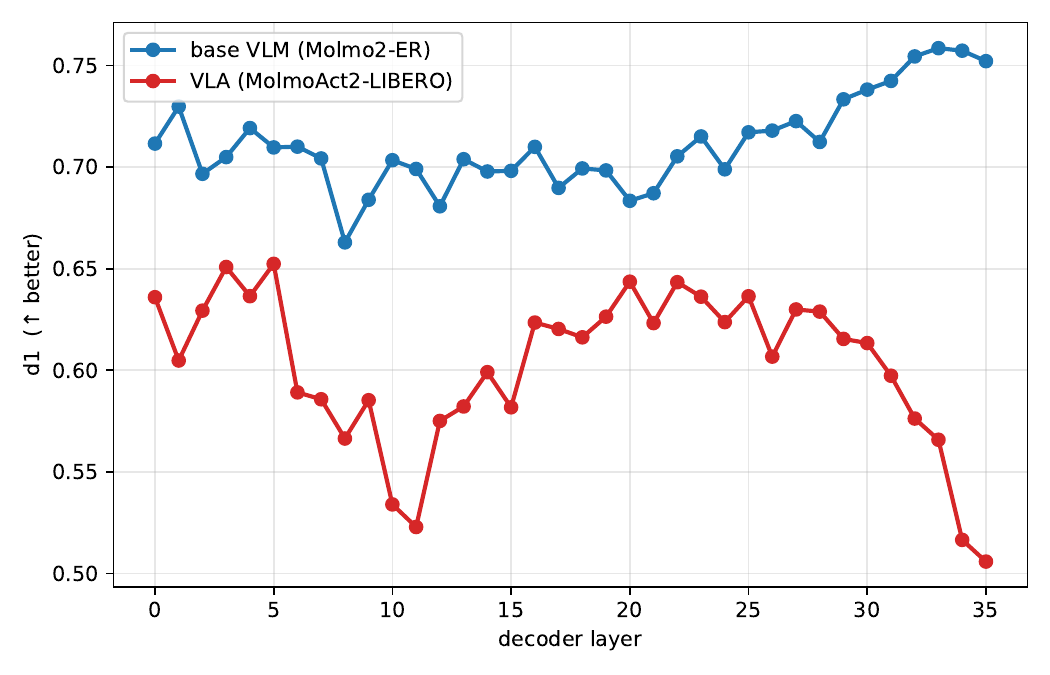}
  \caption{Layerwise depth decodability ($d_1$, higher is better) for Molmo2-ER and MolmoAct2-LIBERO. Both share an early drop and mid-stack recovery, but diverge in the final blocks, where the VLM recovers to its global maximum and the VLA collapses to its global minimum.}
  \label{fig:d1}
\end{figure}

\begin{table}[htbp]
\centering
\caption{Depth decodability ($d_1$) by layer band. Values are means over layers in each band, except the final-layer row.}
\label{tab:bands}
\begin{tabular}{lccc}
\toprule
Layer band & VLM $d_1$ & VLA $d_1$ & $\Delta$ \\
\midrule
Early (L0--8) & 0.705 & 0.617 & 0.089 \\
Mid (L9--27) & 0.701 & 0.606 & 0.095 \\
Late (L28--35) & 0.744 & 0.578 & 0.166 \\
Final layer (L35) & 0.752 & 0.506 & \textbf{0.246} \\
Terminal slope L28$\to$L35 & $+0.040$ & $-0.123$ & -- \\
\bottomrule
\end{tabular}
\end{table}

\paragraph{The Floor and the Cliff.}
As depicted in \cref{fig:d1} and \cref{tab:bands}, the base VLM reproduces the drop-off-to-recovery template reported for adapter-based MLLMs~\cite{wu2026dropoff}: depth is already decodable at L0 ($d_1\approx0.71$), dips to a trough near layer 8 ($d_1\approx0.66$), then recovers to its maximum in the final blocks ($d_1\approx0.76$). This matters as a positive control: before action tuning, the same probing setup recovers a known layerwise pattern and shows that this base VLM normally consolidates depth at the end of the stack. By contrast, the VLA is worse everywhere, with its curve below the VLM's at all 36 layers. The gap is narrowest mid-stack ($\approx0.04\,d_1$ at L20) but never closes. This is the floor.

Furthermore, the degradation is not uniform. Both the VLM and VLA drop early together, then the VLA approaches the base VLM in the middle blocks. The curves diverge sharply at the last 7 layers. Over L28--L35, the base VLM rises by $+0.040$ while the VLA falls by $-0.123$; the VLM peaks in its final layers, whereas the VLA reaches its global minimum there. This is the cliff and the VLA's inversion. The inversion is doubly surprising: transferable geometry is often expected to be most accessible in middle layers, yet the base VLM peaks late (shown first for semantic tasks in Wu et al.~\cite{wu2026dropoff}) and action tuning disproportionately damages precisely those layers where the base model consolidates depth.

\paragraph{Controls.}
The inversion is a property of the hidden states. One probe class, capacity, and training regimen is used across all $2\times36$ model-layer cells, and splitting by rollout prevents leakage of held-out rollouts. The inversion appears under RMSE as well, detailed in the supplementary material. If the DPT head were producing depth largely from its own inductive bias or generic image features, scores should be relatively insensitive to the layer where the head is attached. Instead, the performance of this identical head varies sharply across layers and especially between models: note the $0.25\,d_1$ final-layer gap on identically ordered frames. Together with the ridge corroboration in \cref{sec:decomp}, this strongly rules against a probe-confounded explanation for the pattern. While a probe identifies what is decodable, module-level attribution still remains ambiguous. As articulated by Eq.~\eqref{eq:block}, the residual stream at each layer is a sum of deletable write terms, which we next ablate module-wise and layer-wise.

\subsection{Causal Localization}
\label{sec:causal}
If a specific module's writes interfere with final-layer depth signal, deleting them should produce a more depth-decodable final representation; if the interference is diffuse, no single deletion should stand out. We zero a module's write into the residual stream, either $m^i$ or $a^i$ in Eq.~\eqref{eq:block}, over a window of three consecutive layers, and read depth at the fixed final layer L35 with a freshly trained DPT probe per condition. Twelve non-overlapping windows tile the stack. The design is $2\times2\times12$: MLP or attention, VLA or weight-matched VLM, and window. Additional training and implementation details are provided in the supplementary material.

\begin{table}[htbp]
\centering
\caption{Final-layer depth decodability ($d_1$ at L35) under windowed ablation.}
\label{tab:ablation}
\begin{tabular}{lcccc}
\toprule
Condition & Clean & L33--35 & $\Delta$ late & Max $\Delta$, other windows \\
\midrule
VLA, MLP ablated & 0.506 & \textbf{0.584} & \textbf{+0.078} & +0.050 (L30–32) \\
VLA, attention ablated & 0.506 & 0.537 & +0.031 & +0.033 (L15--17) \\
VLM, MLP ablated & 0.752 & 0.767 & +0.015 & +0.015 (L15--17) \\
VLM, attention ablated & 0.752 & 0.749 & $-0.003$ & +0.010 (L3--5) \\
\bottomrule
\end{tabular}
\end{table}

The symmetry of \cref{tab:ablation} is important. Both module types, both models, and every three-layer window are tested against a shared clean baseline; no design element privileges MLPs, late layers, or the action-trained model. The sweep is therefore hypothesis-neutral at the level of localization: a localized increase is meaningful only if it emerges from the full $2\times2\times12$ comparison.

\begin{figure}[htbp]
  \centering
  \includegraphics[width=.78\linewidth]{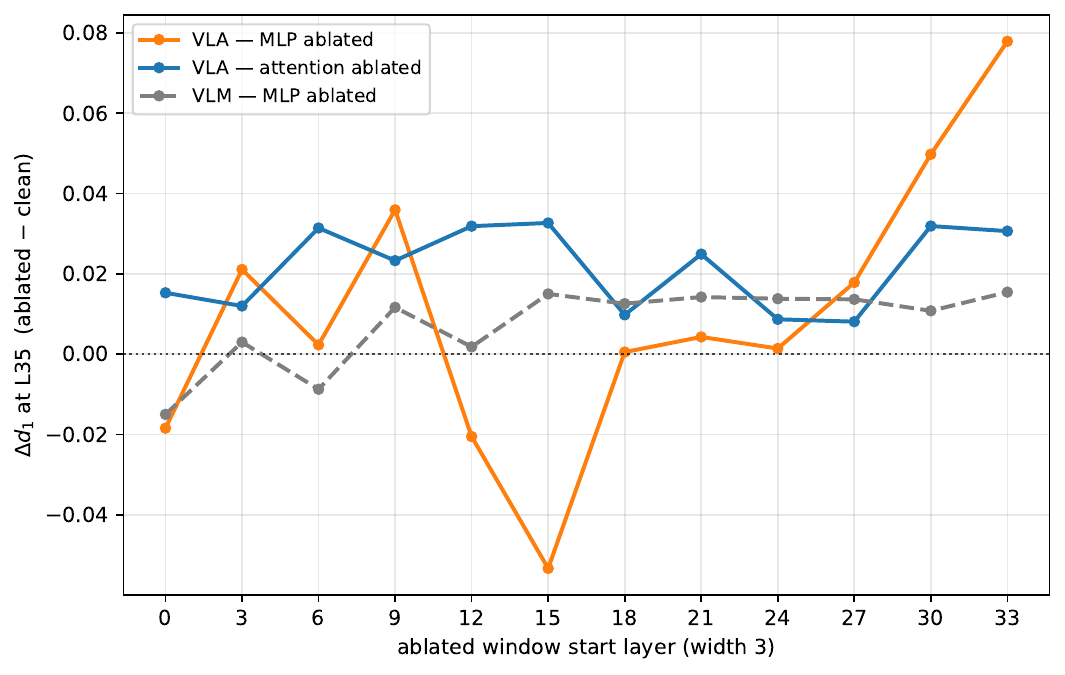}
  \caption{Change in final-layer depth decodability ($\Delta d_1$ at L35) as a function of the ablated three-layer window. Only the VLA's late MLP windows substantially increase final-layer depth decodability, and the increase grows toward the readout.}
  \label{fig:ablation}
\end{figure}

\paragraph{Ablating Late MLP Writes Improves Depth Decodability.}
As plotted in \cref{fig:ablation}, with its final MLP window (L33--35) ablated, the VLA's final-layer depth decodability rises from $d_1=0.506$ to $0.584$, a $+0.078$ improvement. This restores well over half of the $0.123$ cliff measured in \cref{sec:depth}, from a subtractive intervention that deletes writes and adds nothing. The immediately preceding window (L30--32) improves $d_1$ by $+0.050$; every earlier MLP window changes $d_1$ by at most $+0.036$.

Neither control reproduces the magnitude of the VLA MLP decodable depth signal restoration. Attention ablation yields only a smaller, comparatively window-insensitive lift, reaching $+0.031$ in the final window, while the corresponding late-MLP intervention in the base VLM remains within the variation of its sweep at $+0.015$. The full comparison therefore isolates a late, MLP-specific, action-training-specific recovery signature.

The sweep causally localizes the cliff to late VLA MLP computation. Removing the final MLP writes raises $d_1$ from $0.506$ to $0.584$, recovering the majority of the terminal drop, while no attention window or corresponding base-VLM intervention produces comparable recovery. Late MLP writes therefore interfere with final-layer depth decodability in the action-trained model. Because each intervention receives a freshly trained probe, this result establishes recovery of decodability rather than identity of the underlying representational code. This effect concerns final-layer depth decodability, not closed-loop policy quality. The ablation probes are single-seed; consequently, our evidence rests on the structured module-wise, layer-wise, and training-specific dissociation rather than significance of an isolated cell.

\subsection{Where Depth Lives, and What Action-Tuning Does to It}
\label{sec:decomp}
The stream read in \cref{sec:depth} is a shared channel that every block reads from and writes to. We therefore probe the stream against the accumulated MLP writes, the running sum $\sum_{i=0}^{\ell}m^i$ of Eq.~\eqref{eq:block}, hereafter the \emph{MLP deposits} at layer $\ell$.

\begin{figure}[htbp]
  \centering
  \includegraphics[width=\linewidth]{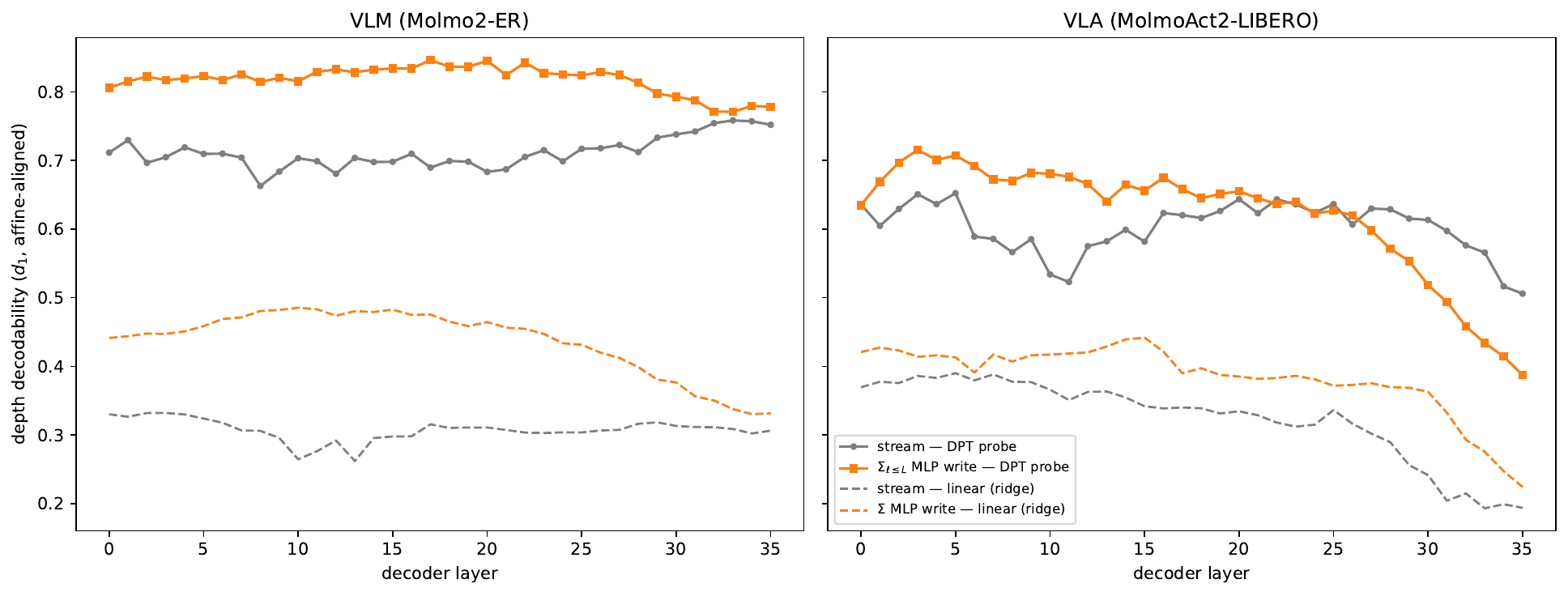}
  \caption{Depth decodability ($d_1$) of the residual stream versus accumulated MLP deposits $\sum_{i=0}^{\ell}m^i$. Solid: DPT probe. Dashed: linear ridge probe. In the VLM, deposits out-decode the stream throughout; in the VLA, deposits collapse over the final blocks.}
  \label{fig:deposits}
\end{figure}

\begin{figure}[t]
  \centering
  \includegraphics[width=\linewidth]{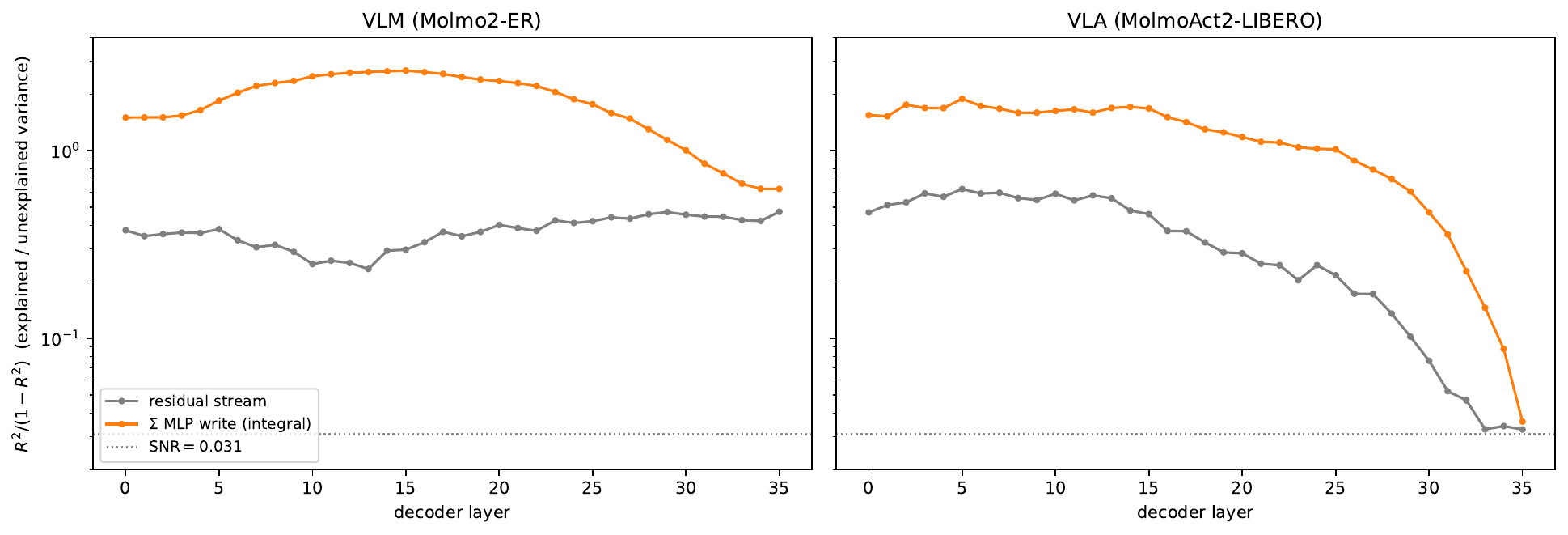}
  \caption{Linear-probe depth SNR, $R^2/(1-R^2)$, for stream and accumulated MLP deposits. In the VLA both collapse over the final layers (SNR=0.03 line for reference) while in the VLM both remain above it.}
  \label{fig:snr}
\end{figure}

\paragraph{Depth Rides the MLP Deposits.}
As shown in \cref{fig:deposits}, in the base VLM, accumulated MLP deposits are more depth-decodable than the stream they write to at every layer: $d_1\approx0.80$--$0.84$ from deposits versus $0.66$--$0.75$ from the stream, converging only in the final blocks as the stream recovers. 
It remains an open question precisely why this gap emerges. We hypothesize that the large magnitude of the initial visual and embedding tokens (relative to early MLP writes) introduces unaligned variance that temporarily drowns out the depth signal in the aggregate stream. We emphasize that this does not imply an isolated MLP write somehow holds more absolute information than the full state. Rather, the data demonstrates that the MLP pathway isolates a cleaner, more easily probe-accessible subspace for geometric features than the aggregate residual stream. A linear ridge probe reproduces, within each model, the same deposits $>$ stream decodability ordering and qualitative trajectory, offset downward as expected from a higher-bias instrument.

Crucially, while the VLM and VLA deposit trajectories closely track each other through the first two-thirds of the stack, they diverge sharply in the final layers. Over these final blocks, the VLA's MLP deposit decodability collapses from $d_1\approx0.68$ down to $0.38$---dropping even below its own residual stream (\cref{fig:deposits}, right). By contrast, the VLM's deposits remain highly decodable until they meet the recovering stream. As shown in \cref{fig:snr}, evaluating linearly recoverable depth confirms this within-model collapse in SNR of our ridge regressor: the VLA's late deposits fall from explaining a majority of depth variance to nearly zero ($SNR \approx 0.03$) by the final layer.

\begin{figure}[htbp]
  \centering
  \includegraphics[width=\linewidth]{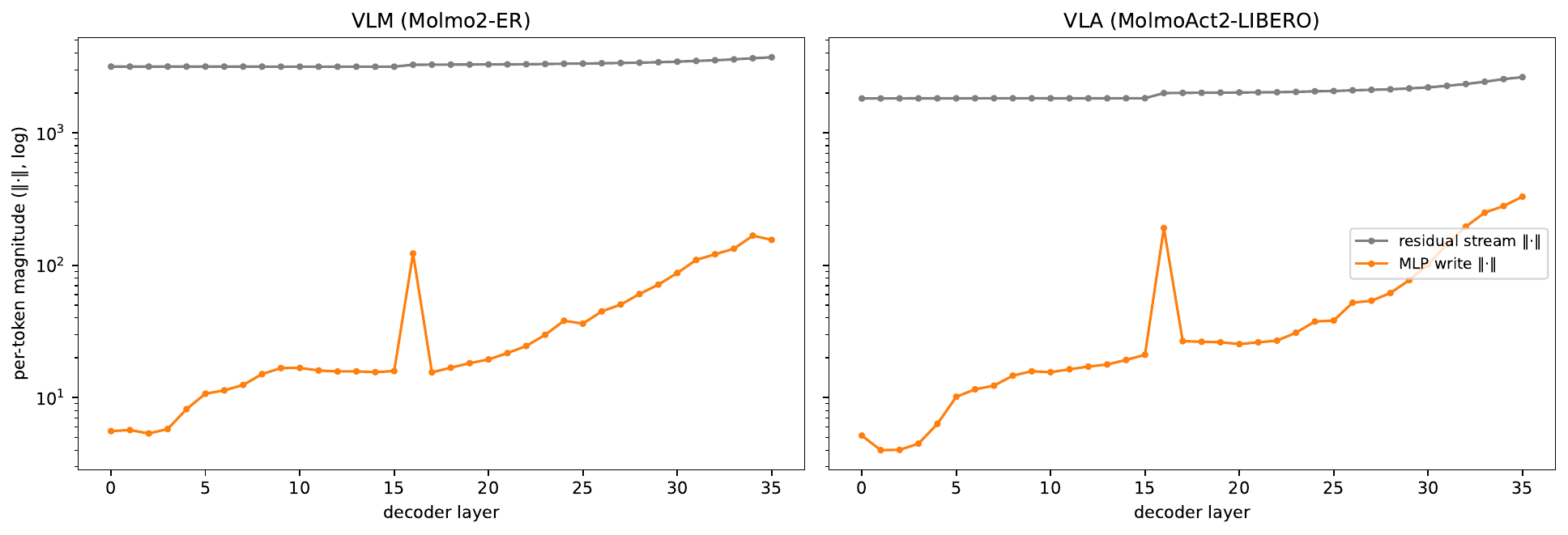}
  \caption{Per-token magnitude of the MLP write versus the residual stream it enters (log axis). The stream is much larger early, compressing to roughly $10\times$ late, so a single write can substantially reshape the stream mainly in the final layers. The MLP write spike at L16 appears to be a massive-activation artifact in both models.}
  \label{fig:mag}
\end{figure}

\paragraph{Why Writes Are the Cleaner Vessel.}
Shown in \cref{fig:mag}, the residual stream carries per-token magnitude $100$--$1000\times$ larger than any single MLP write through the first two-thirds of the stack and slightly over $10\times$ in the final layers. A write is therefore a small residual on a large accumulated vector: its depth-relevant content is a minor fraction of the stream's total variance. This offers an explanation for why depth can be more accessible in accumulated writes than in the shared stream, and why a single late write can exert a larger relative perturbation late in the stack, matching the late-window concentration of the ablation effect in \cref{sec:causal}.

\section{Limitations and Conclusion}
\label{sec:conclusion}

This paper presents a mechanistic study asking how much of one spatiogeometric primitive, depth perception, remains decodable after action post-training, where the reduction emerges across the decoder, and which module-level computations cause the final-layer cliff.

Our study covers one VLM--VLA pair on one dataset: Molmo2-ER and MolmoAct2-LIBERO, evaluated on LIBERO frames. This pairing is deliberate: the complete action post-training pipeline is publicly documented, including its datasets and reimplementation details. The models' shared initialization and weight-matched architecture remove major architectural confounds, while leaving the action-post-training pipeline---including its data and stages---as the bundled treatment under study. Our instruments are also limited to one capacity-matched DPT probe class, corroborated by ridge regression. The target values are determined by Depth-Anything-3 as a pseudo-ground-truth, affine-aligned at evaluation, so the curves measure agreement with a strong monocular estimator rather than metric ground truth. Depth is necessary but not sufficient for spatiogeometric understanding. Finally, our causal claims are representation-level and single-seed; we make no claim that the ablations improve closed-loop policy behavior.

The findings are nonetheless clear. The VLA exhibits worse depth decodability than its base VLM throughout the network, forming a persistent floor, and then collapses across the final blocks where the base VLM instead recovers, forming a terminal cliff. A symmetric full-stack ablation sweep causally localizes the cliff to late MLP computation: deleting those writes recovers the majority of the terminal decodability drop, with an effect specific to that module, those layers, and the action-trained model. Module-level probing explains this localization. In the base VLM, accumulated MLP writes form the most depth-decodable pathway through the decoder; after action post-training, that pathway collapses over the final blocks and its late writes interfere with final-layer depth decodability. Together, these results support a mechanistic account in which action post-training repurposes late MLP computation at the expense of geometric readout.

The promise of a VLA is an embodied policy that inherits a pretrained VLM's understanding of the world. Our diagnosis does not yet provide a training-time remedy, but it separates a persistent cross-layer gap from an additional late-stage cliff that can be partially recovered by deleting the responsible writes. As robot policies increasingly inherit perception from VLMs rather than learn it from scratch, that distinction provides a concrete starting point for repair.

\section*{Acknowledgements}
Our work was fully funded by Reflex, which we thank for their support.

\bibliographystyle{splncs04}
\bibliography{main}

\clearpage
\section*{Supplementary Materials}
\appendix
\renewcommand{\theHsection}{appendix.\Alph{section}}

\section{Depth Probing and Implementation Details}
\label{app:depth}
We train one fresh probe per model-layer cell. The data are LIBERO frames sampled at stride 5 per rollout, resized to 256 pixels, with primary and wrist views. The teacher is DA3MONO-LARGE. The Dense Prediction Transformer probe follows Banani et al. \cite{banani2024probe3d} with an AdaBins/sigmoid depth readout, width 512, kernel size 3, and depth range $[0.001,10]$ at 41.8M trainable parameters. Probe3D's DPT fuses four backbone layers; because our analysis reads a single decoder layer at a time, our single-tap adaptation feeds the one tapped layer into all four DPT inputs, keeping the RefineNet fusion architecture intact while changing only the tap site. Inputs are per-grid z-scored. Optimization uses AdamW, learning rate $5\times10^{-4}$, batch size 8, 20 epochs, and linear warmup over 15\% of steps. A 25\% validation split is held out by rollout, with seed 0.

\begin{table}[htbp]
\centering
\caption{\textbf{Probing-experiment configuration.}}
\label{tab:probe-config}
\small
\begin{tabular*}{0.96\linewidth}{@{\extracolsep{\fill}}>{\raggedright\arraybackslash}p{0.33\linewidth}>{\centering\arraybackslash}p{0.56\linewidth}@{}}
\toprule
\textbf{Configuration} & \textbf{Value} \\
\midrule
\textbf{Models} & Molmo2-ER; MolmoAct2-LIBERO \\
\textbf{Backbone taps} & All 36 decoder layers at visual-token positions \\
\textbf{Data} & LIBERO, rollout-strided, 256p, $n=400$ \\
\textbf{Teacher} & DA3MONO-LARGE \\
\textbf{Probe} & Banani et al. 41.8M parameter DPT head ~\cite{banani2024probe3d} \\
\textbf{Optimizer} & AdamW, $\eta=5\times10^{-4}$, first 15\% linear warmup \\
\textbf{Regimen} & batch size of 8, 20 epochs \\
\textbf{Split} & 25\% hold out over unseen rollouts \\
\textbf{Loss} & Scale-invariant log loss + gradient term \\
\textbf{Evaluation} & best held-out $d_1$; per-image least-squares scale-shift alignment \\
\bottomrule
\end{tabular*}
\end{table}

\begin{table}[p]
\centering
\caption{Per-layer depth decodability, source data for Fig. 3 of the main text}
\label{tab:perlayer}
\scriptsize
\begin{tabular}{rccccc}
\toprule
Layer & VLM $d_1$ & VLM RMSE & VLA $d_1$ & VLA RMSE & $\Delta d_1$ \\
\midrule
0&0.712&0.264&0.636&0.307&0.076\\
1&0.730&0.250&0.605&0.315&0.125\\
2&0.697&0.277&0.629&0.305&0.067\\
3&0.705&0.269&0.651&0.301&0.054\\
4&0.719&0.256&0.636&0.308&0.083\\
5&0.710&0.260&0.652&0.293&0.057\\
6&0.710&0.262&0.589&0.331&0.121\\
7&0.704&0.270&0.586&0.329&0.118\\
8&0.663&0.303&0.566&0.344&0.096\\
9&0.684&0.282&0.585&0.331&0.099\\
10&0.703&0.265&0.534&0.365&0.169\\
11&0.699&0.267&0.523&0.364&0.176\\
12&0.681&0.278&0.575&0.340&0.106\\
13&0.704&0.263&0.582&0.341&0.122\\
14&0.698&0.277&0.599&0.338&0.099\\
15&0.698&0.277&0.582&0.347&0.116\\
16&0.710&0.264&0.623&0.318&0.086\\
17&0.690&0.282&0.620&0.326&0.069\\
18&0.699&0.277&0.616&0.324&0.083\\
19&0.698&0.271&0.626&0.323&0.072\\
20&0.683&0.285&0.644&0.313&0.040\\
21&0.687&0.289&0.623&0.322&0.064\\
22&0.705&0.269&0.643&0.319&0.062\\
23&0.715&0.267&0.636&0.316&0.079\\
24&0.699&0.274&0.624&0.323&0.075\\
25&0.717&0.261&0.636&0.318&0.081\\
26&0.718&0.258&0.607&0.326&0.111\\
27&0.723&0.253&0.630&0.316&0.093\\
28&0.712&0.257&0.629&0.318&0.083\\
29&0.733&0.246&0.615&0.334&0.118\\
30&0.738&0.244&0.613&0.326&0.125\\
31&0.742&0.243&0.597&0.336&0.145\\
32&0.754&0.234&0.576&0.341&0.178\\
33&0.758&0.231&0.566&0.351&0.193\\
34&0.757&0.233&0.517&0.375&0.241\\
35&0.752&0.238&0.506&0.385&0.246\\
\bottomrule
\end{tabular}
\end{table}

\section{Causal-Ablation Details}
\label{app:ablation}
The readout layer is fixed at L35. Twelve non-overlapping three-layer windows tile the stack, starting at layers $0,3,\ldots,33$. For each model, module, and window, the named module's write into the residual stream is zeroed during a forward pass; a clean replica is extracted in the same pass and serves as the shared baseline. The readout probe is identical to ~\cref{app:depth}. Joint ablation of both writes makes the window an identity map and is excluded from the main analysis because it re-reads an earlier layer.

\section{Module-Level Decomposition Details}
\label{app:decomp}
For every decoder block we hook the residual stream and MLP write at visual-token positions in an identical row order across both models. DPT probing of accumulated writes uses the same probe recipe as ~\cref{app:depth}, changing only the tap site from $h^\ell$ to $\sum_{i\le\ell}m^i$. For linear corroboration we fit ridge regressions of per-token log-depth on activations, with columns standardized by train statistics and a relative regularization grid selected by validation $R^2$. We report both $d_1$ and $R^2/(1-R^2)$ on held-out data.

\end{document}